\documentclass[11pt,a4paper]{article}
\usepackage[hyperref]{acl2021}
\usepackage{times}
\usepackage{inconsolata}
\usepackage{latexsym}
\usepackage{amsmath}
\usepackage{array,multirow, graphicx}
\usepackage{booktabs}
\usepackage{xspace}
\usepackage{adjustbox}
\usepackage{comment}
\usepackage{amssymb}
\usepackage{bbding}

\usepackage{cleveref}

\definecolor{darkred}{rgb}{0.55, 0.0, 0.0}
\definecolor{darkgoldenrod}{rgb}{0.72, 0.53, 0.04}
\definecolor{darkspringgreen}{rgb}{0.09, 0.45, 0.27}

\newcolumntype{R}[2]{%
    >{\adjustbox{angle=#1,lap=\width-(#2)}\bgroup}%
    l%
    <{\egroup}%
}

\newcommand{\prem}{$\mathcal{P}$\xspace}
\newcommand{\hypo}{$\mathcal{H}$\xspace}

\usepackage{microtype}

\aclfinalcopy % Uncomment this line for the final submission
\title{AMR4NLI: Interpretable and robust NLI measures from semantic graphs}

\author{
\\
\textbf{Juri Opitz}$^\text{\SixFlowerPetalDotted}$\ \ ~~\textbf{Shira Wein}$^\text{\FourClowerOpen}$\ \ ~~ \textbf{Julius Steen}$^\text{\SixFlowerPetalDotted}$\ \ ~~\textbf{Anette Frank}$^\text{\SixFlowerPetalDotted}$\ \ ~~\textbf{Nathan Schneider}$^\text{\FourClowerOpen}$ \\\\
$^\text{\SixFlowerPetalDotted}$Heidelberg University %Department of Computational Linguistics\\ 
\ \ ~~$^\text{\FourClowerOpen}$Georgetown University \\
\texttt{opitz.sci@gmail.com}~~~ \texttt{\{steen,frank\}@cl.uni-heidelberg.de}\hspace{0.5cm} \\
\hspace{0.5cm}\texttt{\{sw1158,nathan.schneider\}@georgetown.edu}
\\
}

\date{}

\begin{document}
\maketitle
\begin{abstract} 
The task of natural language inference (NLI) asks whether a given premise (expressed in NL) entails a given NL hypothesis. NLI benchmarks contain human ratings of entailment, but the meaning relationships driving these ratings are not formalized. Can the underlying sentence pair relationships be made more explicit in an interpretable yet robust fashion? We compare semantic structures to represent premise and hypothesis, including \textit{sets of contextualized embeddings} and \textit{semantic graphs} (Abstract Meaning Representations), and measure whether the hypothesis is a semantic substructure of the premise, utilizing interpretable metrics. Our evaluation on three English benchmarks finds value in both contextualized embeddings and semantic graphs; moreover, they provide complementary signals, and can be leveraged together in a hybrid model.
\end{abstract}

\section{Introduction}

Natural language inference (NLI) and textual entailment (TE) assess whether a hypothesis (\hypo) is entailed by a premise (\prem). Systems have various interesting applications, e.g., the validation of automatically generated text \cite{holtzman-etal-2018-learning, honovich-etal-2022-true-evaluating}. Recent systems make use of neural networks to encode \hypo and \prem into a vector and thereupon make a prediction \cite{jiang2019evaluating}.  While this can provide strong results when such systems are trained on large-scale training data, the overall decision process is not transparent and may rely more on spurious cues than on informed decisions \cite{poliak-etal-2018-hypothesis}.

\begin{figure}
    \centering
    \includegraphics[width=1.0\linewidth, trim={0 3mm 0 0}]{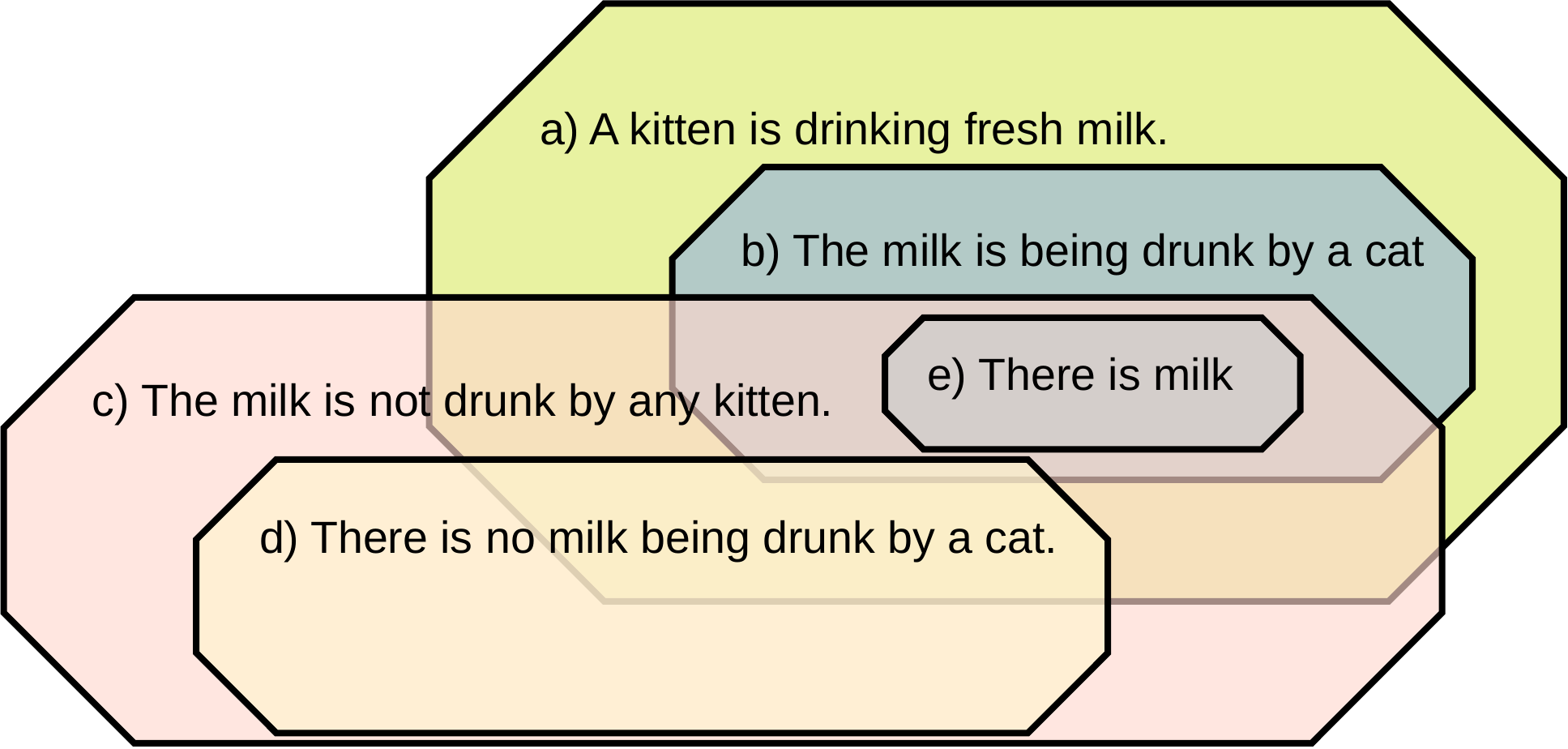}
    \caption{Semantic (sub-)structure analysis shows that 4 of 25 candidate relations are true entailment relations: b) is entailed by a). d) is entailed by c). e) is entailed by a), b), and c). }
    \label{fig:motivation_example}
\end{figure}

We aim to develop more transparent alternatives for NLI prediction, and therefore compare representations and metrics to predict entailment. Figure~\ref{fig:motivation_example} gives an intuition of how 5~different sentences overlap in meaning. Representing each sentence with a semantic structure, we assume that, by and large, the semantic elements of an entailed sentence should be contained within the premise.

These considerations trigger three interesting research questions that we will investigate in this paper: RQ1.\ \textit{How to characterize a semantic structure?} RQ2.\  \textit{How to determine/measure what is a substructure?} RQ3.\  \textit{Is there a suitable and interpretable structure and measure that help to make NLI judgments more robust, or more accurate?}

To assess RQ1, we test three options: token sets, sets of contextualized embeddings, or graph-based meaning representations (MRs). As a meaning representation, we select Abstract Meaning Representation \citep[AMR;][]{banarescu-etal-2013-abstract}, using automatic AMR parses of the NLI sentences. To assess RQ2, we test different types of metrics that are designed or adapted to  measure entailment on the selected structures, inspired from research on, e.g., MT evaluation and MR similarity. One of our key goals is to investigate whether it is possible to accurately capture relevant semantic substructure relationships via meaning representations. Finally, we show that we can positively answer all aspects of RQ3: First, besides their enhanced interpretability, unsupervised semantic graph metrics are more robust and generalize better than fine-tuned BERT. Second, importantly, we show that they are high-precision NLI predictors, a property that we exploit to achieve strong NLI results with a simple decomposable hybrid model built from a fine-tuned BERT on the one hand, and a semantic graph score on the other. Code and data are available at \url{https://github.com/flipz357/AMR4NLI}.

\section{Related work}
\label{sec:related_work}

\paragraph{Textual entailment}  
Automatic approaches for this task date back to, at least, \citet{10.1007/11736790_9}, who introduced a shared task for entailment classification. Since then, we can distinguish many different kinds of systems for addressing the task \cite{androutsopoulos2010survey}, for instance, based on logics \cite{bos-markert-2005-recognising} or string- and tree-similarity \cite{zhang2005paraphrase}, or graph matches of semantic frames and syntax \cite{burchardt2006approaching} that aim in a similar direction as us. Recent releases of large-scale training corpora, such as SNLI \cite{bowman-etal-2015-large}, or MNLI \cite{williams-etal-2018-broad} can be exploited for supervised training of strong classifiers, e.g., by fine-tuning a BERT language model \cite{devlin-etal-2019-bert}. However, trained systems tend to suffer from the `Clever Hans' effect and fall prey to spurious cues \cite{niven-kao-2019-probing, jin2020bertrobust}, such as position \cite{ko-etal-2020-look} or even gender \cite{sharma2021evaluating}. This can lead to undesired and peculiar NLI system behavior. \citet{poliak-etal-2018-hypothesis} show that supervised NLI systems can make many correct predictions solely based on \prem, without even seeing \hypo. In our work, we want to test more transparent ways of rating entailment.

\paragraph{Metrics and meaning representations} In part due to the reduced dependence on spurious cues,  unsupervised/zero-shot metrics are found in evaluation of MT (e.g., BERTscore \cite{bert-score}, BLEURT \cite{sellam-etal-2020-bleurt}), and NLG faithfulness checks \cite{honovich-etal-2022-true-evaluating}. Through the lens of abstract meaning representation \cite{banarescu-etal-2013-abstract}, systems perform explainable sentence similarity \cite{opitz-etal-2021-explainable, opitz-frank-2022-sbert}, NLG evaluation \cite{opitz-frank-2021-towards, manning-schneider-2021-referenceless}, cross-lingual AMR analysis \cite{wein-schneider-2021-classifying,wein-schneider-2022-accounting, wein-etal-2022-effect}, and search \cite{bonial-etal-2020-infoforager, muller2022evaluation, opitz2022smaragd}. \citet{leung-etal-2022-semantic} discuss different use-cases of embedding-based and MR-based metrics.

\section{Method}

\subsection{Underlying research hypotheses}

\paragraph{RH1: Semantic substructure analysis with asymmetric metrics  can predict entailment} We 
aim to study the entailment problem through analysis of semantic structure of \prem and \hypo. To perform such analysis, we need a metric that can measure the degree to which \hypo-structure is contained in the \prem-structure. Therefore, we hypothesize that an \textit{asymmetric metric} is preferable. Note that asymmetric metrics of complex objects like sets or graphs tend to be under-studied in NLP.\footnote{Indeed, most metrics used in NLP are \textit{naturally symmetric} (e.g., cosine distance). Others fuse two asymmetric metrics into, e.g., an F1 score from precision and recall \cite{popovic-2015-chrf,bert-score}. Alternatively, they are inherently asymmetric but enforce symmetry via balancing with an inversely correlated metric, e.g., BLEU \cite{papineni-etal-2002-bleu} focuses on precision but tries to factor in recall via a `brevity penalty'. Even in related cases, where using an asymmetric metric seems intuitive, we find that sometimes symmetric metrics being used instead, e.g., \citet{ribeiro-etal-2022-factgraph} design a baseline for assessing faithfulness of automatically generated summaries with a symmetric F1 score using an AMR metric.}

\paragraph{RH2: Meaning representations are suitable semantic structures} Semantic structures for \prem/\hypo  should (ideally) hold facts that make them true. In this work we explore three options to build such structures for \hypo/\prem: i) the set of text tokens, ii) the set of (contextual) embeddings obtained from them, and iii) graph-structured MRs. It is the latter that we hope will represent the facts best: A token set holds `facts' in their surface form, which can be lossy in morphologically rich languages or with paraphrases. Contextual embedding sets, on the other hand, are powerful meaning representations, but hardly offer interpretability. An MR-structure is semantically more explicit, and is defined to represent a sentence's meaning through  its parts.

\subsection{Implementation}

\paragraph{Preliminaries} 

Let us define a
\begin{equation}
    metric^{\mathcal{D}}_T: \mathcal{D} \times \mathcal{D} \rightarrow [0, 1]
\end{equation}

\noindent
where 1 implies true entailment. With the parameter $\mathcal{D}$ we 
denote the metric domain (i.e., text with $metric^{text}$ or MR with $metric^{graph}$). The type parameter $T$ specifies whether the metric is symmetric ($metric_{sym}$), or asymmetric ($metric_{asym}$).

\subsection{Text metrics: $metric^{text}$}

\paragraph{Token metrics} Given a set of tokens from \hypo and from \prem,  our asymmetric $metric^{text}_{asym}$ calculates a unigram \textit{precision}-score:
\begin{equation}\label{eq:tokprec}
   \text{TokP} =|\mathcal{H}|^{-1} \cdot |toks(\mathcal{H}) \cap toks(\mathcal{P})|,
\end{equation}

\noindent
which is known to be a simple but strong predictor baseline for NLI-related tasks such as faithfulness evaluation in generation \cite{lavie2004significance,banerjee-lavie-2005-meteor,FADAEE18.432} (the most closely related `BLEU-1' is used in many papers to assess system outputs). By switching \hypo and \prem in Eq.\ \ref{eq:tokprec}, we calculate TokR, and based on these a symmetric $metric^{text}_{sym}$ TokS via harmonic mean.

\paragraph{BERTscore \cite{bert-score} is a contextual embedding metric} that calculates a greedy match between BERT embeddings of two texts, in our case: hypothesis $E^{\mathcal{H}}:=embeds(\mathcal{H})$ and premise $E^{\mathcal{P}}:=embeds(\mathcal{P})$. For our asymmetric $metric^{text}_{asym}$, we calculate a precision-based score:
\begin{align}\label{eq:bsprec}
\begin{split}
\text{BertScoP}=|E^{\mathcal{H}}|^{-1}\sum_{e \in E^{\mathcal{H}}} \max_{e' \in E^{\mathcal{P}}} e^Te'.
\end{split}
\end{align}

\noindent
Symmetric $metric^{text}_{sym}$ BertS is calculated as harmonic mean of BertScoP and BertScoR, the latter being obtained by switching $\mathcal{H}$ and $\mathcal{P}$ in Eq.\ \ref{eq:bsprec}.

\subsection{MR Graph metrics: $metric^{graph}$}

We study the following (a)symmetric MR metrics.

\paragraph{GTok} Emulating TokP and TokS, we introduce GTokS and GTokP via Eq.\ \ref{eq:tokprec} applied to two bags of graphs' node- and edge-labels. 

\paragraph{Structural matching with Smatch} \cite{cai-knight-2013-smatch} aligns triples of two graphs for best matching score, and returns precision (SmatchP) and a symmetric F1 score (SmatchS). We use the optimal ILP implementation of \citet{opitz-2023-smatch}.

\paragraph{Contextualized matching with WWLK} aims at a joint and contextualized assessment of node semantics and node semantics informed by neighborhood structures. Therefore, \citet{opitz2021weisfeiler} first iteratively contextualize a vector representation for each node by averaging the embeddings of all nodes in their immediate neighborhood (the iteration count is indicated by K, which we set to 1). The normalized Euclidean distance of the concatenation of these refined vectors defines a cost matrix $C$, where $C_{ij}$ is the distance of nodes $i \in \text{\prem}$, $j \in \text{\hypo}$. The AMR similarity score is derived by solving a transportation problem: $WWLK = 1 - \min_{F} \sum_{i}\sum_j  F_{ij} C_{ij}$ where $F_{ij}$ is the flow between nodes $i, j$. \citeauthor{opitz2021weisfeiler}\ constrain $\sum_{j} F_{*j} = 1 / |\text{\prem}|$ and $\sum_{i} F_{i*} = 1 / |\text{\hypo}|$. We call this symmetric setting WWLK\textbf{S}. We additionally propose an asymmetric sub-graph matching score WWLK\textbf{P} where we let $\sum_{j} F_{*j} \leq 1$ instead. 

The most reduced version, which deletes all structural information from the graphs, is achieved by setting $k=0$, which we denote as N(ode)Mover(P$\vert$S) score, analogously to the popular word mover's score \cite{kusner2015word}.

\subsection{Hybrid model}

Our decomposable hybrid model takes the prediction of a text metric, and the prediction of a graph metric, and returns an aggregate score. Such a metric can provide an interesting balance between a score grounded in a linguistic interpretation, and a score obtained from strong language models. If the two scores are both useful \textit{and} complementary, we may even hope for a rise in overall results. To test such a scenario we will combine the best performing $metric_{graph}$ with the best performing $metric_{text}$ via a simple sum ($\alpha=0.5$):
\begin{equation}
\label{eq:hybridweighting}
     \alpha \cdot metric^{graph} + (1 - \alpha) metric^{text}.
\end{equation}

\section{Evaluation setup}

\paragraph{Data sets} We employ five standard sentence-level data sets: i) \textbf{SICK (test)} by \citet{marelli-etal-2014-sick} and \textbf{SNLI (dev \& test)} by \citet{bowman-etal-2015-large}, as well as iii) \textbf{MNLI (matched \& mismatched)} by \citet{williams-etal-2018-broad}. Mismatched (henceforth referred to as MNLI-mi) can be understood as a supposedly more challenging data set since it contains entailment problems from a different domain than the training data, allowing a more robust generalization assessment of trained models. By contrast, in MNLI-ma(tched) the domain of the testing data matches that of the training data. For each data set, we map the three NLI labels to a binary TE classification setting, by merging \textit{contradiction} and \textit{neutral} to the \textit{non-entailed} class.\footnote{Same as in \citet{uhrig-etal-2021-translate}, we use the T5-based off-the-shelf parser from \href{https://github.com/bjascob/amrlib}{amrlib} for projecting AMR structures.}

\paragraph{Evaluation metric} 
We expect predictions to correlate with the probability of entailment, i.e.,
\begin{equation}
\label{eq:objective}
    metric^{\mathcal{D}}_T(x,y) \uparrow \implies P(x~\text{entails}~y) \uparrow,
\end{equation}
\noindent
where $\uparrow$ means `rising' (i.e., if the metric assigns a higher score, we expect greater probability of entailment). The NLI `gold probability' labels are approximated as binary human majority labels. To circumvent a threshold search and obtain a meaningful evaluation score for comparing our metrics, we follow the advice of \citet{honovich-etal-2022-true-evaluating}, who evaluate metrics for zero-shot faithfulness evaluation of automatic summarization systems, using the \textit{Area Under Curve} (AUC) metric. The AUC score equals the probability that given randomly drawn instances ($\mathcal{P}, \mathcal{H}$, entailed) and ($\mathcal{P}', \mathcal{H}'$, non-entailed) the entailed instance receives a higher score. To rank metrics, we calculate two averages: AVG$^{all}$ averages the scores over all data sets, while AVG$^{nli}$ excludes SICK.\footnote{SICK contains entailment labels but not the direction of entailment and thus we do not include it in AVG$^{nli}$.}

\paragraph{Trained (upper-bound)} We use a BERT trained on 500k SNLI examples.\footnote{\url{https://huggingface.co/textattack/bert-base-uncased-snli}} It predicts an entailment probability from a vector representation generated by a transformer model.

\begin{table*}[t]
    \centering
    \scalebox{0.82}{
    \begin{tabular}{llrrrrr|rr}
    \toprule
      $\mathcal{D}$(omain) & $metric$  &  SICK & SNLI-dev & SNLI-test & MNLI-ma & MNLI-mi & AVG$^{all}$ & AVG$^{nli}$ \\
       \cmidrule{1-9}
        \parbox[t]{2mm}{\multirow{4}{*}{\rotatebox[origin=c]{90}{text}}} & TokS & 72.1 & 64.2 & 64.6 & 66.7 & 68.7 & 67.2 & 66.0 \\
         & TokP &  74.7 & 70.0 & 70.6 & 68.2 & 70.3 & 70.8 & 69.8 \\
         & BertScoS & 79.8 & 66.7 & 66.2 & 68.4 & 71.6 & 70.5 & 68.2 \\
         & BertScoP & \textbf{82.0} & 74.5 & 74.0 & \textbf{74.5} & \textbf{77.5} & 76.5 & 75.1 \\
         \midrule
        \parbox[t]{2mm}{\multirow{8}{*}{\rotatebox[origin=c]{90}{AMR graph}}} & GTokS & 78.2 & 63.2 & 62.6 & 66.4 & 68.5 & 67.8 & 65.2 \\
         & GTokP  & 81.0 & 75.1 & 74.7 & 71.1 & 72.6 & 74.9 & 73.4 \\
         & NMoverS & 77.7 & 65.8 & 64.9 & 66.7 & 68.5 & 68.7 & 66.5 \\
         & NMoverP & 79.4 & 77.9 & 77.2 & 72.9 & 74.8 & 76.5 & 75.7 \\
         & SmatchS & 76.3 & 63.3 & 62.3 & 65.7 & 67.6 & 67.0 & 64.7 \\
         & SmatchP & 79.2 & 72.3 & 71.6 & 70.0 & 71.9 & 73.0 & 71.4 \\
         & WWLKS & 77.2 & 66.4 & 65.6 & 65.7 & 67.5 & 68.5 & 66.3 \\
         & WWLKP & 79.3 & \textbf{78.0} & \textbf{77.3} & 71.9 & 73.8 & 76.1 & 75.3 \\
         \midrule
         \midrule
        text & trainBERT & 81.0 & 88.8 & 88.2 & 71.5 & 72.0 & 80.3 & 80.1 \\
        hybrid & trainBERT + WWLKP  &  \textbf{85.9} & \textbf{91.0} & \textbf{90.4} & \textbf{77.9} & \textbf{78.9} & \textbf{84.8} & \textbf{84.5} \\
         \bottomrule
         \bottomrule
    \end{tabular}}
    \caption{Overall AUC results on five data sets. The last two rows involve a trained component.}
    \label{tab:mainres}
\end{table*}

\begin{table*}[ht]
    \centering
    \scalebox{0.83}{
    \begin{tabular}{llrrrrrrrr|rr}
    \toprule
    & & \multicolumn{8}{c}{AVG Accuracy scores} \\
       $\mathcal{D}$(omain) &  $metric$ &  1\% & 2\% & 3\% & 4\%& 5\% & 7\% & 10\% & 15\% & AVG$^{all}$ & AVG$^{nli}$ \\
       \cmidrule{1-12}
        \parbox[t]{2mm}{\multirow{2}{*}{\rotatebox[origin=c]{90}{text}}} & TokP & 88.4 & 87.1 & 81.0 & 74.4 & 72.8 & 71.4 & 68.3 & 64.2 & 76.0 & 77.3 \\
         & BertScoP & 74.5 & 74.0 & 73.3 & 73.9 & 73.9 & 73.0 & 72.0 & 69.4 & 73.0 & 73.8 \\
         \midrule
        \parbox[t]{2mm}{\multirow{4}{*}{\rotatebox[origin=c]{90}{AMR graph}}} & GTokP & 86.5 & 86.5 & 87.1 & 88.0 & 87.7 & 86.1 & 80.4 & 73.6 & 84.5 & 88.4 \\
       &  NMoverP &85.3 & 84.5 & 85.0 & 85.2 & 86.2 & 84.7 & 82.4 & 74.2 & 83.4 & 89.6 \\
         & SmatchP & 90.0 & 89.1 & 88.4 & 85.2 & 81.9 & 77.9 & 74.2 & 68.3 & 81.9 & 83.8 \\
       &  WWLKP & \textbf{97.3} & \textbf{96.8} & \textbf{96.1} & \textbf{95.0} & \textbf{93.8} & 88.4 & 82.4 & 74.8 & 90.6 & 90.7 \\
       \midrule
       \midrule
      text & trainBERT & 84.5 & 84.0 & 82.9 & 81.5 & 80.6 & 79.0 & 76.8 & 73.2 & 80.3 & 81.9 \\
    hybrid  & trainBERT + WWLKP & 96.7 & 95.7 & 94.3 & 93.4 & 92.5 & \textbf{90.2} & \textbf{86.7} & \textbf{82.2} & \textbf{91.5} & \textbf{92.9} \\
       \bottomrule
    \end{tabular}}
    \caption{Precision assessment. We select p\% of a metric's highest predictions and check the ratio of true entailment.}
    \label{tab:confident}
    \vspace{-3mm}
\end{table*}

\section{Results}

\subsection{Main insights} Main insights can be inferred from Table \ref{tab:mainres}.  On all data sets, and overall on average, \textbf{asymmetric metrics substantially outperform symmetric metrics}. Sometimes they improve results by up to ten AUC points over their symmetric counterparts (e.g., NMoverS vs.\ NMoverP, +9.2). Comparing token sets, embedding sets and graphs, we find that both embedding set and graph prove advantageous: NMoverP achieves slightly better results than BertScoP, which has been \textit{pre-trained} on large data. \textit{Fine-tuned} BERT outperforms the tested unsupervised metrics when test data is in-domain (see SNLI results), but falls short at generalization. However, our \textbf{simple hybrid model can inform the output with sub-graph overlap and yields a strong boost outperforming all unsupervised and even trained metrics by a large margin (+4.5 points)}.

\subsection{Analysis}

\begin{table}[]
    \centering
    \scalebox{0.82}{
    \begin{tabular}{l||r|r|r|r|r}
    \toprule
       training & \multicolumn{2}{c|}{no} & yes & no & no/yes \\
       
       domain & text & \multicolumn{2}{c|}{text embedding} & AMR & hybrid \\
       metric  & TokP & BScoP & BERT &  WWLKP & +BERT \\
       \midrule
       AUC & 71.0 & 71.4 & 76.2 &  77.7 & 83.1  \\
        AUC $\Delta$ & \textbf{\textcolor{darkspringgreen}{+0.4}} & \textbf{\textcolor{darkgoldenrod}{-3.6}} & \textbf{\textcolor{darkred}{-12.0}} & \textbf{\textcolor{darkspringgreen}{+0.4}} & \textbf{\textcolor{darkgoldenrod}{-7.3}} \\
        \bottomrule
    \end{tabular}}
    \caption{Evaluation on 3,261 \textit{hard} SNLI-test examples. AUC $\Delta$: observed change in performance (cf.\ Table \ref{tab:mainres}).}
    \label{tab:hardexamples}
\end{table}

\paragraph{Advantage of AMR and AMR metrics: high precision} For each metric, we retrieve the p\% most probable predictions, and calculate their accuracy. Results, averaged over all data sets, are displayed in Table \ref{tab:confident}. In high \% levels, MR metrics outperform BertScoP by almost 20 points (e.g., BertScoP vs.\ WWLKP: +17.6 points), and even the fine-tuned BERT is strongly outperformed. Therefore, we can attribute the surprisingly strong performance of the graph metrics (and the hybrid model) to its potential for delivering high scores in which we can trust -- if it determines that the semantic graph of \hypo is (largely) a subgraph of \prem, true entailment is most likely (in Appendix \ref{sec:entailmentexamples}, we show two examples). 

\paragraph{Advantage of untrained (AMR) metrics: better robustness} We check the robustness of our diverse NLI metrics on a controlled subset of 3,261 SNLI testing examples by \citet{gururangan-etal-2018-annotation}, who removed examples that show spurious biases and/or annotation artifacts. Results in Table \ref{tab:hardexamples} show a catastrophic performance drop by trained BERT (\textminus{}12.0 points), while untrained metrics such as TokP and WWLKP remain unaffected (+0.4 points) and WWLKP now even outperforms the SNLI-trained BERT model. Lastly, we see that the hybrid model can (partially) mitigate the drop introduced by its trained component (\textminus{}7.3 points).

\paragraph{Discussion: graph metrics struggle with recall, and other limitations} The MR metrics struggle with recall since they have problems to cope with MRs that strongly differ structurally, but not (much) semantically, which is a known issue \cite{opitz2021weisfeiler}. An example from our data is the following: In \textit{The man rages}, \textit{man} is the \textit{arg0} of rage, while in the entailed sentence \textit{A person is angry}, \textit{person} is the \textit{arg1} of \textit{angry}, yielding large structural dissimilarity of MR graphs (SmatchP=0.0). In future work we aim to explore and improve this issue, such that we are able to identify that the experiencer of \textit{angry} is strongly related to the \textit{agent} of \textit{rage}. 

Potentially unrelated to the recall problem, other issues may hamper AMR usage for NLI, e.g., inconsistent copula modeling \cite{venant-lareau-2023-predicates}, or parsing errors: even though parsers tend to provide high-quality output structures, they can still suffer from significant flaws \cite{opitz-frank-2022-better}, and thus their improvement may positively affect AMR4NLI performance. 

\begin{figure}
    \centering
    \includegraphics[width=\linewidth]{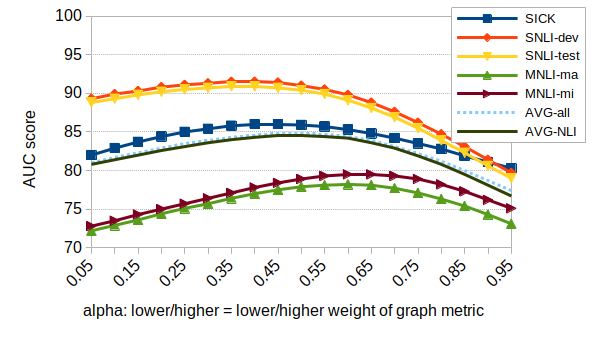}
    \caption{Balancing the hybrid text-graph metric.}
    \label{fig:alphaweighting}
\end{figure}

\paragraph{Weights in hybrid model} Recall that we can use $\alpha$ in Eq.\ \ref{eq:hybridweighting} to weigh two metrics. We inspect different $\alpha$ in Figure \ref{fig:alphaweighting} for fusing trainBERT (text) and WWLKP (graph, $\alpha \geq 0.5$: graph metric is weighted higher). While a balance ($\alpha \approx 0.5$) overall seems effective, SNLI profits if the text metric has more influence, and MNLI profits if the graph metric dominates. Finally, again we see more stable performance of graph metrics overall (converging AUC with high $\alpha$ vs.\ diverging AUC with low $\alpha$).

\section{Conclusion}

We find that metrics defined on advanced semantic representations are useful predictors of entailment. This is especially true for metrics performing asymmetric measurements on graph-structured meaning representations and sets of contextualized embeddings. Interestingly, meaning representation-based metrics  offer advantages over strong embedding-based metrics beyond just interpretability: while showing similar performance as BERTscore, they are more robust than fine-tuned BERT \textit{and} offer high-precision predictions. With this, we show that linguistic and neural representations can complement each other in a hybrid model, leading to substantial improvement over both untrained and trained neural approaches.

\section*{Acknowledgments}

We thank anonymous reviewers for their feedback. This work is partially supported by a Clare Boothe Luce Scholarship.

\bibliographystyle{acl_natbib}
\bibliography{acl2021}

\begin{thebibliography}{45}
\expandafter\ifx\csname natexlab\endcsname\relax\def\natexlab#1{#1}\fi

\bibitem[{Androutsopoulos and Malakasiotis(2010)}]{androutsopoulos2010survey}
Ion Androutsopoulos and Prodromos Malakasiotis. 2010.
\newblock A survey of paraphrasing and textual entailment methods.
\newblock \emph{Journal of Artificial Intelligence Research}, 38:135--187.

\bibitem[{Banarescu et~al.(2013)Banarescu, Bonial, Cai, Georgescu, Griffitt,
  Hermjakob, Knight, Koehn, Palmer, and
  Schneider}]{banarescu-etal-2013-abstract}
Laura Banarescu, Claire Bonial, Shu Cai, Madalina Georgescu, Kira Griffitt, Ulf
  Hermjakob, Kevin Knight, Philipp Koehn, Martha Palmer, and Nathan Schneider.
  2013.
\newblock \href {https://www.aclweb.org/anthology/W13-2322} {Abstract meaning
  representation for sembanking}.
\newblock In \emph{Proceedings of the 7th Linguistic Annotation Workshop and
  Interoperability with Discourse}, pages 178--186, Sofia, Bulgaria.
  Association for Computational Linguistics.

\bibitem[{Banerjee and Lavie(2005)}]{banerjee-lavie-2005-meteor}
Satanjeev Banerjee and Alon Lavie. 2005.
\newblock \href {https://www.aclweb.org/anthology/W05-0909} {{METEOR}: An
  automatic metric for {MT} evaluation with improved correlation with human
  judgments}.
\newblock In \emph{Proceedings of the {ACL} Workshop on Intrinsic and Extrinsic
  Evaluation Measures for Machine Translation and/or Summarization}, pages
  65--72, Ann Arbor, Michigan.

\bibitem[{Bonial et~al.(2020)Bonial, Lukin, Doughty, Hill, and
  Voss}]{bonial-etal-2020-infoforager}
Claire Bonial, Stephanie~M. Lukin, David Doughty, Steven Hill, and Clare Voss.
  2020.
\newblock \href {https://www.aclweb.org/anthology/2020.dmr-1.7}
  {{I}nfo{F}orager: Leveraging semantic search with {AMR} for {COVID}-19
  research}.
\newblock In \emph{Proceedings of the Second International Workshop on
  Designing Meaning Representations}, pages 67--77, Barcelona Spain (online).
  Association for Computational Linguistics.

\bibitem[{Bos and Markert(2005)}]{bos-markert-2005-recognising}
Johan Bos and Katja Markert. 2005.
\newblock \href {https://aclanthology.org/H05-1079} {Recognising textual
  entailment with logical inference}.
\newblock In \emph{Proceedings of Human Language Technology Conference and
  Conference on Empirical Methods in Natural Language Processing}, pages
  628--635, Vancouver, British Columbia, Canada. Association for Computational
  Linguistics.

\bibitem[{Bowman et~al.(2015)Bowman, Angeli, Potts, and
  Manning}]{bowman-etal-2015-large}
Samuel~R. Bowman, Gabor Angeli, Christopher Potts, and Christopher~D. Manning.
  2015.
\newblock \href {https://doi.org/10.18653/v1/D15-1075} {A large annotated
  corpus for learning natural language inference}.
\newblock In \emph{Proceedings of the 2015 Conference on Empirical Methods in
  Natural Language Processing}, pages 632--642, Lisbon, Portugal. Association
  for Computational Linguistics.

\bibitem[{Burchardt and Frank(2006)}]{burchardt2006approaching}
Aljoscha Burchardt and Anette Frank. 2006.
\newblock Approaching textual entailment with lfg and framenet frames.
\newblock In \emph{Proc. of the Second PASCAL RTE Challenge Workshop.[-]}.

\bibitem[{Cai and Knight(2013)}]{cai-knight-2013-smatch}
Shu Cai and Kevin Knight. 2013.
\newblock \href {https://www.aclweb.org/anthology/P13-2131} {{S}match: an
  evaluation metric for semantic feature structures}.
\newblock In \emph{Proceedings of the 51st Annual Meeting of the Association
  for Computational Linguistics (Volume 2: Short Papers)}, pages 748--752,
  Sofia, Bulgaria. Association for Computational Linguistics.

\bibitem[{Dagan et~al.(2006)Dagan, Glickman, and Magnini}]{10.1007/11736790_9}
Ido Dagan, Oren Glickman, and Bernardo Magnini. 2006.
\newblock The pascal recognising textual entailment challenge.
\newblock In \emph{Machine Learning Challenges. Evaluating Predictive
  Uncertainty, Visual Object Classification, and Recognising Tectual
  Entailment}, pages 177--190, Berlin, Heidelberg. Springer Berlin Heidelberg.

\bibitem[{Devlin et~al.(2019)Devlin, Chang, Lee, and
  Toutanova}]{devlin-etal-2019-bert}
Jacob Devlin, Ming-Wei Chang, Kenton Lee, and Kristina Toutanova. 2019.
\newblock \href {https://doi.org/10.18653/v1/N19-1423} {{BERT}: Pre-training of
  deep bidirectional transformers for language understanding}.
\newblock In \emph{Proceedings of the 2019 Conference of the North {A}merican
  Chapter of the Association for Computational Linguistics: Human Language
  Technologies, Volume 1 (Long and Short Papers)}, pages 4171--4186,
  Minneapolis, Minnesota. Association for Computational Linguistics.

\bibitem[{Fadaee et~al.(2018)Fadaee, Bisazza, and Monz}]{FADAEE18.432}
Marzieh Fadaee, Arianna Bisazza, and Christof Monz. 2018.
\newblock {Examining the Tip of the Iceberg: A Data Set for Idiom Translation}.
\newblock In \emph{Proceedings of the Eleventh International Conference on
  Language Resources and Evaluation (LREC 2018)}, Miyazaki, Japan. European
  Language Resources Association (ELRA).

\bibitem[{Gururangan et~al.(2018)Gururangan, Swayamdipta, Levy, Schwartz,
  Bowman, and Smith}]{gururangan-etal-2018-annotation}
Suchin Gururangan, Swabha Swayamdipta, Omer Levy, Roy Schwartz, Samuel Bowman,
  and Noah~A. Smith. 2018.
\newblock \href {https://doi.org/10.18653/v1/N18-2017} {Annotation artifacts in
  natural language inference data}.
\newblock In \emph{Proceedings of the 2018 Conference of the North {A}merican
  Chapter of the Association for Computational Linguistics: Human Language
  Technologies, Volume 2 (Short Papers)}, pages 107--112, New Orleans,
  Louisiana. Association for Computational Linguistics.

\bibitem[{Holtzman et~al.(2018)Holtzman, Buys, Forbes, Bosselut, Golub, and
  Choi}]{holtzman-etal-2018-learning}
Ari Holtzman, Jan Buys, Maxwell Forbes, Antoine Bosselut, David Golub, and
  Yejin Choi. 2018.
\newblock \href {https://doi.org/10.18653/v1/P18-1152} {Learning to write with
  cooperative discriminators}.
\newblock In \emph{Proceedings of the 56th Annual Meeting of the Association
  for Computational Linguistics (Volume 1: Long Papers)}, pages 1638--1649,
  Melbourne, Australia. Association for Computational Linguistics.

\bibitem[{Honovich et~al.(2022)Honovich, Aharoni, Herzig, Taitelbaum,
  Kukliansy, Cohen, Scialom, Szpektor, Hassidim, and
  Matias}]{honovich-etal-2022-true-evaluating}
Or~Honovich, Roee Aharoni, Jonathan Herzig, Hagai Taitelbaum, Doron Kukliansy,
  Vered Cohen, Thomas Scialom, Idan Szpektor, Avinatan Hassidim, and Yossi
  Matias. 2022.
\newblock {TRUE}: Re-evaluating factual consistency evaluation.
\newblock In \emph{Proceedings of the 2022 Conference of the North American
  Chapter of the Association for Computational Linguistics: Human Language
  Technologies}, pages 3905--3920, Seattle, United States. Association for
  Computational Linguistics.

\bibitem[{Jiang and de~Marneffe(2019)}]{jiang2019evaluating}
Nanjiang Jiang and Marie-Catherine de~Marneffe. 2019.
\newblock Evaluating bert for natural language inference: A case study on the
  commitmentbank.
\newblock In \emph{Proceedings of the 2019 conference on empirical methods in
  natural language processing and the 9th international joint conference on
  natural language processing (EMNLP-IJCNLP)}, pages 6086--6091.

\bibitem[{Jin et~al.(2020)Jin, Jin, Zhou, and Szolovits}]{jin2020bertrobust}
Di~Jin, Zhijing Jin, Joey~Tianyi Zhou, and Peter Szolovits. 2020.
\newblock Is bert really robust? a strong baseline for natural language attack
  on text classification and entailment.
\newblock In \emph{Proceedings of the AAAI conference on artificial
  intelligence}, volume~34, pages 8018--8025.

\bibitem[{Ko et~al.(2020)Ko, Lee, Kim, Kim, and Kang}]{ko-etal-2020-look}
Miyoung Ko, Jinhyuk Lee, Hyunjae Kim, Gangwoo Kim, and Jaewoo Kang. 2020.
\newblock \href {https://doi.org/10.18653/v1/2020.emnlp-main.84} {Look at the
  first sentence: Position bias in question answering}.
\newblock In \emph{Proceedings of the 2020 Conference on Empirical Methods in
  Natural Language Processing (EMNLP)}, pages 1109--1121, Online. Association
  for Computational Linguistics.

\bibitem[{Kusner et~al.(2015)Kusner, Sun, Kolkin, and
  Weinberger}]{kusner2015word}
Matt Kusner, Yu~Sun, Nicholas Kolkin, and Kilian Weinberger. 2015.
\newblock From word embeddings to document distances.
\newblock In \emph{International conference on machine learning}, pages
  957--966. PMLR.

\bibitem[{Lavie et~al.(2004)Lavie, Sagae, and
  Jayaraman}]{lavie2004significance}
Alon Lavie, Kenji Sagae, and Shyamsundar Jayaraman. 2004.
\newblock The significance of recall in automatic metrics for mt evaluation.
\newblock In \emph{Machine Translation: From Real Users to Research: 6th
  Conference of the Association for Machine Translation in the Americas, AMTA
  2004, Washington, DC, USA, September 28-October 2, 2004. Proceedings 6},
  pages 134--143. Springer.

\bibitem[{Leung et~al.(2022)Leung, Wein, and
  Schneider}]{leung-etal-2022-semantic}
Wai~Ching Leung, Shira Wein, and Nathan Schneider. 2022.
\newblock \href {https://aclanthology.org/2022.gem-1.8} {Semantic similarity as
  a window into vector- and graph-based metrics}.
\newblock In \emph{Proceedings of the 2nd Workshop on Natural Language
  Generation, Evaluation, and Metrics (GEM)}, pages 106--115, Abu Dhabi, United
  Arab Emirates (Hybrid). Association for Computational Linguistics.

\bibitem[{Manning and Schneider(2021)}]{manning-schneider-2021-referenceless}
Emma Manning and Nathan Schneider. 2021.
\newblock \href {https://aclanthology.org/2021.eval4nlp-1.12} {Referenceless
  parsing-based evaluation of {AMR}-to-{E}nglish generation}.
\newblock In \emph{Proceedings of the 2nd Workshop on Evaluation and Comparison
  of NLP Systems}, pages 114--122, Punta Cana, Dominican Republic. Association
  for Computational Linguistics.

\bibitem[{Marelli et~al.(2014)Marelli, Menini, Baroni, Bentivogli, Bernardi,
  and Zamparelli}]{marelli-etal-2014-sick}
Marco Marelli, Stefano Menini, Marco Baroni, Luisa Bentivogli, Raffaella
  Bernardi, and Roberto Zamparelli. 2014.
\newblock \href
  {http://www.lrec-conf.org/proceedings/lrec2014/pdf/363_Paper.pdf} {A {SICK}
  cure for the evaluation of compositional distributional semantic models}.
\newblock In \emph{Proceedings of the Ninth International Conference on
  Language Resources and Evaluation ({LREC}-2014)}, pages 216--223, Reykjavik,
  Iceland. European Languages Resources Association (ELRA).

\bibitem[{M{\"u}ller and Kuwertz(2022)}]{muller2022evaluation}
Almuth M{\"u}ller and Achim Kuwertz. 2022.
\newblock Evaluation of a semantic search approach based on amr for information
  retrieval in image exploitation.
\newblock In \emph{2022 Sensor Data Fusion: Trends, Solutions, Applications
  (SDF)}, pages 1--6. IEEE.

\bibitem[{Niven and Kao(2019)}]{niven-kao-2019-probing}
Timothy Niven and Hung-Yu Kao. 2019.
\newblock \href {https://doi.org/10.18653/v1/P19-1459} {Probing neural network
  comprehension of natural language arguments}.
\newblock In \emph{Proceedings of the 57th Annual Meeting of the Association
  for Computational Linguistics}, pages 4658--4664, Florence, Italy.
  Association for Computational Linguistics.

\bibitem[{Opitz(2023)}]{opitz-2023-smatch}
Juri Opitz. 2023.
\newblock \href {https://aclanthology.org/2023.findings-eacl.118} {{SMATCH}++:
  Standardized and extended evaluation of semantic graphs}.
\newblock In \emph{Findings of the Association for Computational Linguistics:
  EACL 2023}, pages 1595--1607, Dubrovnik, Croatia. Association for
  Computational Linguistics.

\bibitem[{Opitz et~al.(2021{\natexlab{a}})Opitz, Daza, and
  Frank}]{opitz2021weisfeiler}
Juri Opitz, Angel Daza, and Anette Frank. 2021{\natexlab{a}}.
\newblock \href {https://doi.org/10.1162/tacl_a_00435} {{Weisfeiler-Leman in
  the Bamboo: Novel AMR Graph Metrics and a Benchmark for AMR Graph
  Similarity}}.
\newblock \emph{Transactions of the Association for Computational Linguistics},
  9:1425--1441.

\bibitem[{Opitz and Frank(2021)}]{opitz-frank-2021-towards}
Juri Opitz and Anette Frank. 2021.
\newblock \href {https://aclanthology.org/2021.eacl-main.129} {Towards a
  decomposable metric for explainable evaluation of text generation from
  {AMR}}.
\newblock In \emph{Proceedings of the 16th Conference of the European Chapter
  of the Association for Computational Linguistics: Main Volume}, pages
  1504--1518, Online. Association for Computational Linguistics.

\bibitem[{Opitz and Frank(2022{\natexlab{a}})}]{opitz-frank-2022-better}
Juri Opitz and Anette Frank. 2022{\natexlab{a}}.
\newblock \href {https://doi.org/10.18653/v1/2022.eval4nlp-1.4} {Better
  {S}match = better parser? {AMR} evaluation is not so simple anymore}.
\newblock In \emph{Proceedings of the 3rd Workshop on Evaluation and Comparison
  of NLP Systems}, pages 32--43, Online. Association for Computational
  Linguistics.

\bibitem[{Opitz and Frank(2022{\natexlab{b}})}]{opitz-frank-2022-sbert}
Juri Opitz and Anette Frank. 2022{\natexlab{b}}.
\newblock \href {https://aclanthology.org/2022.aacl-main.48} {{SBERT} studies
  meaning representations: Decomposing sentence embeddings into explainable
  semantic features}.
\newblock In \emph{Proceedings of the 2nd Conference of the Asia-Pacific
  Chapter of the Association for Computational Linguistics and the 12th
  International Joint Conference on Natural Language Processing}, pages
  625--638, Online only. Association for Computational Linguistics.

\bibitem[{Opitz et~al.(2021{\natexlab{b}})Opitz, Heinisch, Wiesenbach, Cimiano,
  and Frank}]{opitz-etal-2021-explainable}
Juri Opitz, Philipp Heinisch, Philipp Wiesenbach, Philipp Cimiano, and Anette
  Frank. 2021{\natexlab{b}}.
\newblock \href {https://doi.org/10.18653/v1/2021.argmining-1.3} {Explainable
  unsupervised argument similarity rating with {A}bstract {M}eaning
  {R}epresentation and conclusion generation}.
\newblock In \emph{Proceedings of the 8th Workshop on Argument Mining}, pages
  24--35, Punta Cana, Dominican Republic. Association for Computational
  Linguistics.

\bibitem[{Opitz et~al.(2022)Opitz, Meier, and Frank}]{opitz2022smaragd}
Juri Opitz, Philipp Meier, and Anette Frank. 2022.
\newblock Smaragd: Synthesized smatch for accurate and rapid amr graph
  distance.
\newblock \emph{arXiv preprint arXiv:2203.13226}.

\bibitem[{Papineni et~al.(2002)Papineni, Roukos, Ward, and
  Zhu}]{papineni-etal-2002-bleu}
Kishore Papineni, Salim Roukos, Todd Ward, and Wei-Jing Zhu. 2002.
\newblock \href {https://doi.org/10.3115/1073083.1073135} {{B}leu: a method for
  automatic evaluation of machine translation}.
\newblock In \emph{Proceedings of the 40th Annual Meeting of the Association
  for Computational Linguistics}, pages 311--318, Philadelphia, Pennsylvania,
  USA. Association for Computational Linguistics.

\bibitem[{Poliak et~al.(2018)Poliak, Naradowsky, Haldar, Rudinger, and
  Van~Durme}]{poliak-etal-2018-hypothesis}
Adam Poliak, Jason Naradowsky, Aparajita Haldar, Rachel Rudinger, and Benjamin
  Van~Durme. 2018.
\newblock \href {https://doi.org/10.18653/v1/S18-2023} {Hypothesis only
  baselines in natural language inference}.
\newblock In \emph{Proceedings of the Seventh Joint Conference on Lexical and
  Computational Semantics}, pages 180--191, New Orleans, Louisiana. Association
  for Computational Linguistics.

\bibitem[{Popovi{\'c}(2015)}]{popovic-2015-chrf}
Maja Popovi{\'c}. 2015.
\newblock \href {https://doi.org/10.18653/v1/W15-3049} {chr{F}: character
  n-gram f-score for automatic {MT} evaluation}.
\newblock In \emph{Proceedings of the Tenth Workshop on Statistical Machine
  Translation}, pages 392--395, Lisbon, Portugal. Association for Computational
  Linguistics.

\bibitem[{Ribeiro et~al.(2022)Ribeiro, Liu, Gurevych, Dreyer, and
  Bansal}]{ribeiro-etal-2022-factgraph}
Leonardo F.~R. Ribeiro, Mengwen Liu, Iryna Gurevych, Markus Dreyer, and Mohit
  Bansal. 2022.
\newblock \href {https://doi.org/10.18653/v1/2022.naacl-main.236}
  {{F}act{G}raph: Evaluating factuality in summarization with semantic graph
  representations}.
\newblock In \emph{Proceedings of the 2022 Conference of the North American
  Chapter of the Association for Computational Linguistics: Human Language
  Technologies}, pages 3238--3253, Seattle, United States. Association for
  Computational Linguistics.

\bibitem[{Sellam et~al.(2020)Sellam, Das, and Parikh}]{sellam-etal-2020-bleurt}
Thibault Sellam, Dipanjan Das, and Ankur Parikh. 2020.
\newblock \href {https://doi.org/10.18653/v1/2020.acl-main.704} {{BLEURT}:
  Learning robust metrics for text generation}.
\newblock In \emph{Proceedings of the 58th Annual Meeting of the Association
  for Computational Linguistics}, pages 7881--7892, Online. Association for
  Computational Linguistics.

\bibitem[{Sharma et~al.(2021)Sharma, Dey, and Sinha}]{sharma2021evaluating}
Shanya Sharma, Manan Dey, and Koustuv Sinha. 2021.
\newblock Evaluating gender bias in natural language inference.
\newblock \emph{arXiv preprint arXiv:2105.05541}.

\bibitem[{Uhrig et~al.(2021)Uhrig, Garcia, Opitz, and
  Frank}]{uhrig-etal-2021-translate}
Sarah Uhrig, Yoalli Garcia, Juri Opitz, and Anette Frank. 2021.
\newblock \href {https://doi.org/10.18653/v1/2021.iwpt-1.6} {Translate, then
  parse! a strong baseline for cross-lingual {AMR} parsing}.
\newblock In \emph{Proceedings of the 17th International Conference on Parsing
  Technologies and the IWPT 2021 Shared Task on Parsing into Enhanced Universal
  Dependencies (IWPT 2021)}, pages 58--64, Online. Association for
  Computational Linguistics.

\bibitem[{Venant and Lareau(2023)}]{venant-lareau-2023-predicates}
Antoine Venant and Fran{\c{c}}ois Lareau. 2023.
\newblock \href {https://aclanthology.org/2023.depling-1.4} {Predicates and
  entities in {A}bstract {M}eaning {R}epresentation}.
\newblock In \emph{Proceedings of the Seventh International Conference on
  Dependency Linguistics (Depling, GURT/SyntaxFest 2023)}, pages 32--41,
  Washington, D.C. Association for Computational Linguistics.

\bibitem[{Wein et~al.(2022)Wein, Leung, Mu, and
  Schneider}]{wein-etal-2022-effect}
Shira Wein, Wai~Ching Leung, Yifu Mu, and Nathan Schneider. 2022.
\newblock Effect of source language on {AMR} structure.
\newblock In \emph{Proceedings of The 16th Linguistic Annotation Workshop
  (LAW)}, Marseille, France. European Language Resources Association (ELRA).

\bibitem[{Wein and Schneider(2021)}]{wein-schneider-2021-classifying}
Shira Wein and Nathan Schneider. 2021.
\newblock \href {https://doi.org/10.18653/v1/2021.law-1.6} {Classifying
  divergences in cross-lingual {AMR} pairs}.
\newblock In \emph{Proceedings of the Joint 15th Linguistic Annotation Workshop
  (LAW) and 3rd Designing Meaning Representations (DMR) Workshop}, pages
  56--65, Punta Cana, Dominican Republic. Association for Computational
  Linguistics.

\bibitem[{Wein and Schneider(2022)}]{wein-schneider-2022-accounting}
Shira Wein and Nathan Schneider. 2022.
\newblock \href {https://aclanthology.org/2022.coling-1.336} {Accounting for
  language effect in the evaluation of cross-lingual {AMR} parsers}.
\newblock In \emph{Proceedings of the 29th International Conference on
  Computational Linguistics}, pages 3824--3834, Gyeongju, Republic of Korea.
  International Committee on Computational Linguistics.

\bibitem[{Williams et~al.(2018)Williams, Nangia, and
  Bowman}]{williams-etal-2018-broad}
Adina Williams, Nikita Nangia, and Samuel Bowman. 2018.
\newblock \href {https://doi.org/10.18653/v1/N18-1101} {A broad-coverage
  challenge corpus for sentence understanding through inference}.
\newblock In \emph{Proceedings of the 2018 Conference of the North {A}merican
  Chapter of the Association for Computational Linguistics: Human Language
  Technologies, Volume 1 (Long Papers)}, pages 1112--1122, New Orleans,
  Louisiana. Association for Computational Linguistics.

\bibitem[{Zhang et~al.(2020)Zhang, Kishore, Wu, Weinberger, and
  Artzi}]{bert-score}
Tianyi Zhang, Varsha Kishore, Felix Wu, Kilian~Q. Weinberger, and Yoav Artzi.
  2020.
\newblock \href {https://openreview.net/forum?id=SkeHuCVFDr} {{BERTScore:
  Evaluating Text Generation with BERT}}.
\newblock In \emph{International Conference on Learning Representations}.

\bibitem[{Zhang and Patrick(2005)}]{zhang2005paraphrase}
Yitao Zhang and Jon Patrick. 2005.
\newblock Paraphrase identification by text canonicalization.
\newblock In \emph{Proceedings of the Australasian Language Technology Workshop
  2005}, pages 160--166.

\end{thebibliography}

\clearpage\newpage
\appendix
\section{Appendix}
\label{sec:entailmentexamples}

\begin{figure}[hb]
    \centering
    \includegraphics[width=1.9\linewidth]{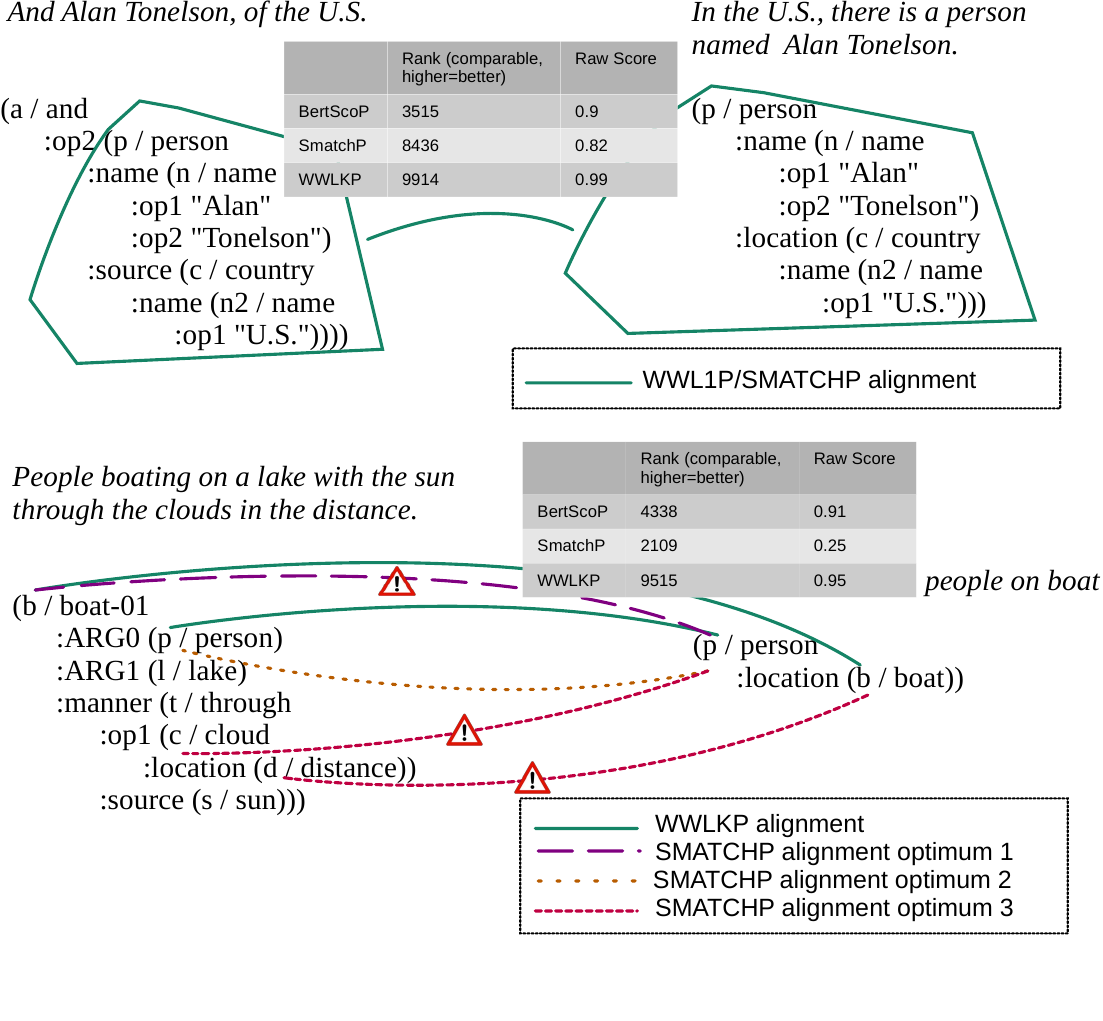}
    \caption{Two example ratings assessing true entailment: The first shows how MR can define a useful semantic set, the second shows that sometimes embedding-based graph metrics, such as WWLKP, are needed to assess the subgraph properly (in this example, SmatchP provides semantically meaningless alignments and a score that is too low.)}
    \label{fig:my_label}
\end{figure}

\end{document}